\documentclass[10pt,twocolumn,letterpaper]{article}

\usepackage{cvpr}              

\usepackage{graphicx}
\usepackage{amsmath}
\usepackage{amssymb}
\usepackage{booktabs}
\usepackage{mathrsfs}

\usepackage[pagebackref,breaklinks,colorlinks]{hyperref}

\usepackage[capitalize]{cleveref}
\crefname{section}{Sec.}{Secs.}
\Crefname{section}{Section}{Sections}
\Crefname{table}{Table}{Tables}
\crefname{table}{Tab.}{Tabs.}

\usepackage{newfloat}
\usepackage{listings}
\usepackage{algorithm}
\usepackage{algorithmic}
\usepackage{multirow}
\usepackage{amssymb}
\usepackage{amsmath}

\def\cvprPaperID{2088} 
\def\confName{CVPR}
\def\confYear{2022}

\usepackage{algorithm}
\usepackage{algorithmic}

\usepackage{booktabs}
\usepackage[accsupp]{axessibility} 

\usepackage{array}
\newcommand{\PreserveBackslash}[1]{\let\temp=\\#1\let\\=\temp}
\newcolumntype{C}[1]{>{\PreserveBackslash\centering}p{#1}}
\newcolumntype{R}[1]{>{\PreserveBackslash\raggedleft}p{#1}}
\newcolumntype{L}[1]{>{\PreserveBackslash\raggedright}p{#1}}

\begin{document}

\title{DRINet++: Efficient Voxel-as-point Point Cloud Segmentation}

\author{
	\begin{tabular}{ p{2.8cm}<{\centering} p{2.0cm}<{\centering} p{2.6cm}<{\centering} p{2.5cm}<{\centering} p{2.4cm}<{\centering} p{2.5cm}<{\centering}}
Maosheng Ye\textsuperscript{1$\ddagger$*} & Rui Wan\textsuperscript{2*} & Shuangjie Xu\textsuperscript{1$\ddagger$} & Tongyi Cao\textsuperscript{2}  & Qifeng Chen\textsuperscript{1} 
\end{tabular}\\
\textsuperscript{1}Hong Kong University of Science and Technology  \quad  \textsuperscript{2}DEEPROUTE.AI\\
{\tt\small {myeag, sxubj}@connect.ust.hk \quad \{ruiwan, tongyicao\}@deeproute.ai \quad cqf@ust.hk}
}

\maketitle

\begin{abstract}
Recently, many approaches have been proposed through single or multiple representations to improve the performance of point cloud semantic segmentation. However, these works do not maintain a good balance among performance, efficiency, and memory consumption. To address these issues, we propose DRINet++ that extends DRINet by enhancing the sparsity and geometric properties of a point cloud with a voxel-as-point principle. To improve efficiency and performance, DRINet++ mainly consists of two modules: Sparse Feature Encoder and Sparse Geometry Feature Enhancement. The Sparse Feature Encoder extracts the local context information for each point, and the Sparse Geometry Feature Enhancement enhances the geometric properties of a sparse point cloud via multi-scale sparse projection and attentive multi-scale fusion. In addition, we propose deep sparse supervision in the training phase to help convergence and alleviate the memory consumption problem. Our DRINet++ achieves state-of-the-art outdoor point cloud segmentation on both SemanticKITTI and Nuscenes datasets while running significantly faster and consuming less memory.   
\end{abstract}

\section{Introduction}

\let\thefootnote\relax\footnotetext{\textsuperscript{$\ddagger$}Part of the work was done during an internship at DEEPROUTE.AI.}
\let\thefootnote\relax\footnotetext{\textsuperscript{$*$}Equal contributions.}

\begin{figure}
\vspace{-10px}
    \centering
    \includegraphics[width=1\linewidth]{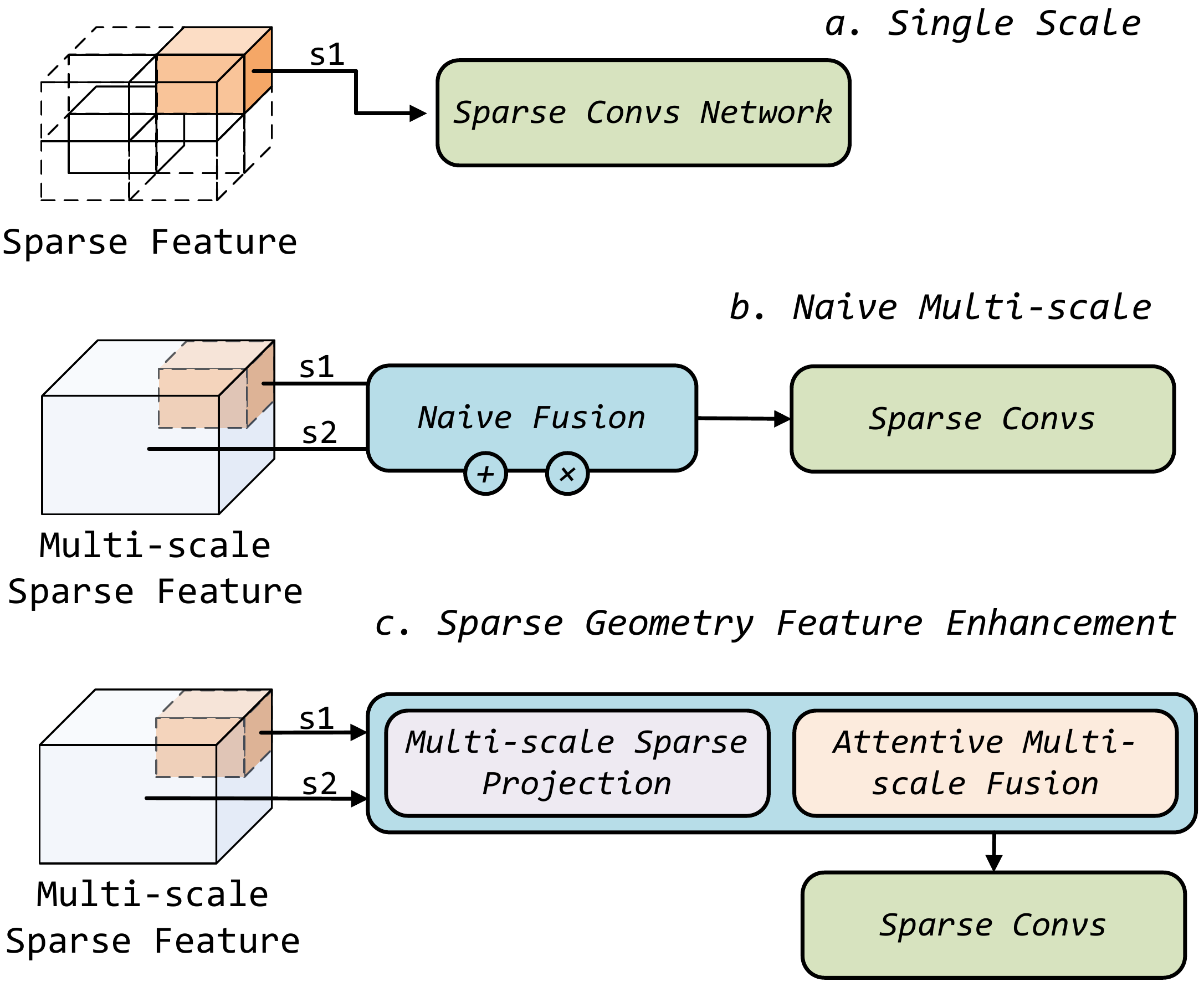}
    \caption{Comparison between two common ways to deal with sparse features and our proposed method.
    }
    \label{fig:some_results}
    \vspace{-5px}
    \end{figure}
    
\noindent Large-scale outdoor point cloud segmentation has been a crucial task for autonomous driving systems and has demanding requirements for efficiency, performance, and memory consumption.
For point cloud segmentation, PointNet~\cite{qi2017pointnet} and PointNet++~\cite{qiPointNetDeepHierarchical2017a} are the pioneering works that directly operate on a point cloud to learn pointwise features by utilizing the key geometric properties (e.g., permutation invariant, Euclidean distances) of a point cloud. However, it is hard to apply these approaches in outdoor scenarios due to high memory consumption and low runtime efficiency. RandLA~\cite{hu2020randla} applies a random sampling strategy to reduce the number of points to improve efficiency, which leads to some information loss. With the popularity of sparse convolutions~\cite{Graham20173D,yan2018second}, there has been some progress in utilizing the sparse voxel-based representation (e.g., AF2S3Net~\cite{cheng20212} and Cylinder3D~\cite{zhu2021cylindrical}), which is a representation that preserves the 3D Euclidean space. Compared with point-based representations, the merits of the sparse voxel-based representation lie in the effectiveness and efficiency in quickly expanding the receptive fields based on the sparsity of a point cloud (another key property). Furthermore, the sparse voxel-based representation, which aggregates point features within the local neighborhood, can significantly reduce memory usage. Also, traditional or current popular convolutional neural networks (CNNs) can be directly applied to extract better context information.

Recently, several works recognize the limitation of a single representation and explore richer information by synergizing multiple representations. PVCNN~\cite{liu2019point} fuses point-based and voxel-based representations with MLP layers and dense 3D convolution layers but does not take point cloud sparsity into consideration. SPVCNN~\cite{tang2020searching}, TPCN~\cite{ye2021tpcn} and DRINet~\cite{ye2021drinet} design the sparse convolution layers and pointwise operational layers to fuse features considering sparsity and geometry. Furthermore, RPVNet~\cite{xu2021rpvnet} combines the range-view, point, and voxel representations for point cloud segmentation.
The general framework of current multi-representation learning is to utilize sparse convolutions for locality and sparsity, and pointwise operations for geometry learning, aiming to combine sparsity and geometry for a better balance between performance and efficiency. While these approaches lead to some performance improvements, 
the extra computation cost and memory usage brought by the additional point representation cannot be ignored considering $100$K points.
Meanwhile, according to the experimental results by these approaches~\cite{ye2021drinet, tang2020searching}, the voxel-based representation is still the dominant one, with which these methods do achieve decent performance. 
Note that a voxel can be a ``\textbf{super point}'' as the abstraction of a set of points within a cube, and the number of voxels can be much less than the number of points as shown in Fig.~\ref{fig:voxel_number}. Therefore, we are raising a question: 
Is it enough to only keep voxel-based representation and treat the voxels as points for efficiency, and apply these pointwise operations to the ``\textbf{super points}'' for better performance? Inspired by these observations, we propose DRINet++ to explore extra geometric properties based on a single sparse voxel-based representation. We extend the architecture from DRINet~\cite{ye2021drinet} 
for its merits in iterative learning, and redesign the pointwise operations based on the principle of ``\textbf{voxels as points}''. Compared with these multi-representation methods, our DRINet++ incorporates sparsity and geometric properties of a point cloud in a unified representation without introducing extra computation cost from multi-representation fusion.

Our DRINet++ extends the design of DRINet~\cite{ye2021drinet} and mainly contains two modules to iteratively enhance feature learning: Sparse Feature Encoder (SFE) and Sparse Geometric Feature Enhancement (SGFE). Each module takes the output of the other module as input to fully explore the sparsity and geometric properties of a point cloud at a low computation cost and memory usage. In SGFE (shown in Fig.~\ref{fig:some_results}), we propose a novel multi-scale sparse projection layer for hierarchical geometry learning and attentive multi-scale fusion for multi-scale feature selection. Apart from that, we apply deep sparse supervision compared with the most common dense manner to alleviate the pressure of memory consumption. Compared with DRINet~\cite{ye2021drinet}, our DRINet++ treats the voxels as points to alleviate the memory and efficiency problem. 

Our contributions are summarized as follows:
\begin{itemize}
\item We propose a new lightweight network architecture called DRINet++ based on the voxel-as-point principle to fully exploit the sparsity and geometry properties of a point cloud. The Multi-scale Sparse Projection and the Attentive Multi-scale Fusion in the Sparse Geometric Feature Enhancement are proposed to enhance geometric feature learning.
\item Deep Sparse Supervision is proposed as a training strategy in a deep and sparse fashion to reduce the memory cost and improve the performance.
\item We evaluate our proposed approach on large-scale outdoor scenario datasets including SemanticKITTI~\cite{behley2019semantickitti} and nuScenes-lidarseg~\cite{caesar2020nuscenes} to demonstrate the effectiveness of our method. We achieve state-of-the-art performance on both datasets with an average running time of $59ms$ per frame with an Nvidia RTX 2080Ti GPU.
\end{itemize}

\section{Related Work}
\textbf{Indoor Point Cloud Segmentation.} The point cloud from a indoor scene often has closely positioned points with a small range. The existing indoor point cloud segmentation approaches can be classified according to their model representations. For point-based approaches, PointNet~\cite{qi2017pointnet}, PointNet++~\cite{qi2017pointnet}, and their related works~\cite{li2018pointcnn,wu2019pointconv,zhang2019linked,liu2020closer,pham2019jsis3d,qi20173d} based on a similar architecture are popular models in this task. Most of these works explore the local neighborhood context while preserving the inherent geometry of a point cloud. They use different grouping and permutation invariant operations to promote performance. The other mainstream methods~\cite{su2015multi, maturana2015voxnet, liu2019point,kumawat2019lp,le2018pointgrid} follow the volumetric representation by partitioning the space as discrete pixels/voxels and then apply 2D/3D CNN architecture to the regular representation. Graph-based approaches for point cloud learning~\cite{wang2018local, zhang2018graph,te2018rgcnn,wang2019graph, wang2019dynamic,xu2020grid} are also popular due to the nature of graph to deal with unorderedness and the capability to model the relationship among points. Currently, with the popularity of transformers, some works~\cite{zhao2020point, guo2020pct} achieve state-of-the-art performance in indoor point cloud learning by introducing transformer-based architectures. Although a number of novel architectures have been proposed to improve point cloud learning, some of them fail to generalize to outdoor scenarios with thousands of hundreds of points. 

\begin{figure*}
\vspace{-20px}
    \centering
    \includegraphics[width=17.5cm]{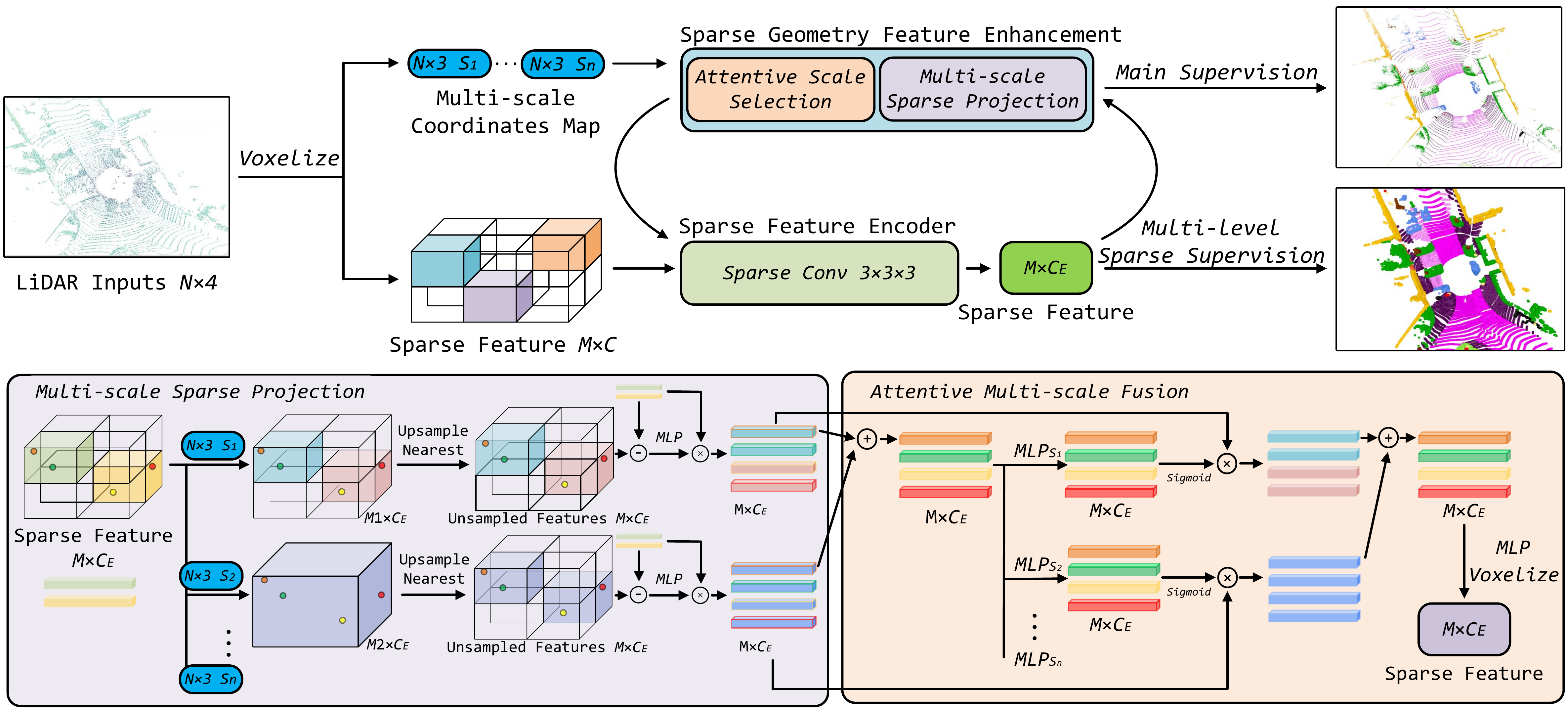}
    \caption{The overall structure of our DRINet++. In the top half of the figure, LiDAR input is firstly voxelized as sparse features. Then the Sparse Feature Encoder utilizes sparse convolutions to process the sparse features. Furthermore, Sparse Geometry Feature Enhancement will enhance the features by Multi-scale Sparse Projection and Attentive Multi-scale Fusion layer to generate the input of Sparse Feature Encoder at the next stage. Sparse supervision will be attached to the output of the sparse feature encoder as an auxiliary loss. The bottom line describes the details about Multi-scale Sparse Projection and Attentive Multi-scale Fusion. {$N$} is the number of points, {$M_{i}$} is the number of voxels for $i$-th scale, $C_E$ is the channel dimension.}
    \label{fig:network}
\vspace{-15px}
\end{figure*}

\noindent\textbf{Outdoor Point Cloud Segmentation.} Compared with indoor point cloud segmentation, the sparsity and larger number of points pose great challenges for existing approaches. Point-based methods such as KPConv~\cite{thomas2019kpconv} and RandLA~\cite{hu2020randla} extend the architecture of PointNet~\cite{qi2017pointnet} or PointNet++~\cite{qiPointNetDeepHierarchical2017a} and adopt sampling strategies to alleviate these problems but lead to extra information loss. KPConv~\cite{thomas2019kpconv} introduces the kernel point selection process to generate high-quality sampling points. Range-view-based approaches~\cite{wu2018squeezeseg, xu2020squeezesegv3,cortinhal2020salsanext} project the point cloud into range views or spherical representations and apply efficient CNN architectures. However, the range view cannot maintain the metric space and introduces distortions, which potentially leads to performance degradation. Some other approaches~\cite{zhu2021cylindrical,zhang2020polarnet,choy20194d,cheng20212,yan2020sparse} quantize a point cloud into some pre-defined space or representations (e.g., polar grids, 2D grids, and sparse 3D grids) and then apply regular convolution neural networks or sparse convolutions~\cite{Graham20173D,yan2018second} to achieve the balance between efficiency and performance. A line of of works integrates the multiple representations, including range views, voxel representations, and point representation, to exploit the potential of different representations deeply \cite{tang2020searching, xu2021rpvnet, ye2021drinet, liu2019point}. These works utilize different architectures for different representations and propose various fusion strategies and show strong performance gain compared to single-representation-based methods, at the cost of extra running time.


\noindent\textbf{Image Segmentation to Point Cloud Segmentation.} The fully convolutional network (FCN)~\cite{long2015fully} is one of the pioneering works for image segmentation with deep learning. Based on FCN and existing prevalent CNN architecture, DeepLab~\cite{chen2017deeplab}, PSPNet~\cite{zhao2017pyramid} and their following works~\cite{yang2018denseaspp, chen2017rethinking} are proposed with multi-scale or multiple dilation rate strategies to explore more hierarchical local context information. Furtherly, HRNet~\cite{sun2019deep} fuses different resolution heatmap in a single framework and keeps the high resolution to improve the performance. Considering the great process achieved in image segmentation, lots of works~\cite{ye2021drinet, rangenet++, unal2021improving, zhang2020polarnet, zhu2021cylindrical} have applied these tricks including hierarchy learning, attention mechanism, or backbones into point cloud segmentation. Some works~\cite{zhu2021cylindrical, choy20194d} are built on U-net~\cite{ronneberger2015u} with sparse convolution acceleration.

\section{Approach}
The overall network architecture of our DRINet++ is based on DRINet~\cite{ye2020hvnet} that propagates features in an iterative manner, but only consists of two different modules: 1) Sparse Feature Encoder 2) Sparse Geometry Feature Enhancement. Sparse Feature Encoder serves as a basic block for fast local context aggregation. Sparse Geometry Feature Enhancement, which takes the output of Sparse Feature Encoder as input, enhances the geometric information by multi-scale sparse projection and attentive multi-scale fusion layer. Both modules interact a proposed voxel-as-point fashion to reduce computation cost and improve runtime efficiency. Compared with previous multi-representation methods~\cite{ye2021drinet, tang2020searching, liu2019point}, we treat the voxels as points and apply geometric operations to mimic the previous pointwise feature learning. Moreover, we propose to apply Deep Sparse Supervision at the voxel level to alleviate the memory issues resulted from dense supervision in sec.~\ref{sec:dss}. Fig.~\ref{fig:network} demonstrates the overall framework of our proposed approach. Limitations are discussed in the supplementary materials.


\subsection{Voxels As Points}
\begin{figure}[!h]
    \vspace{-5px}
    \centering
    \includegraphics[width=8cm]{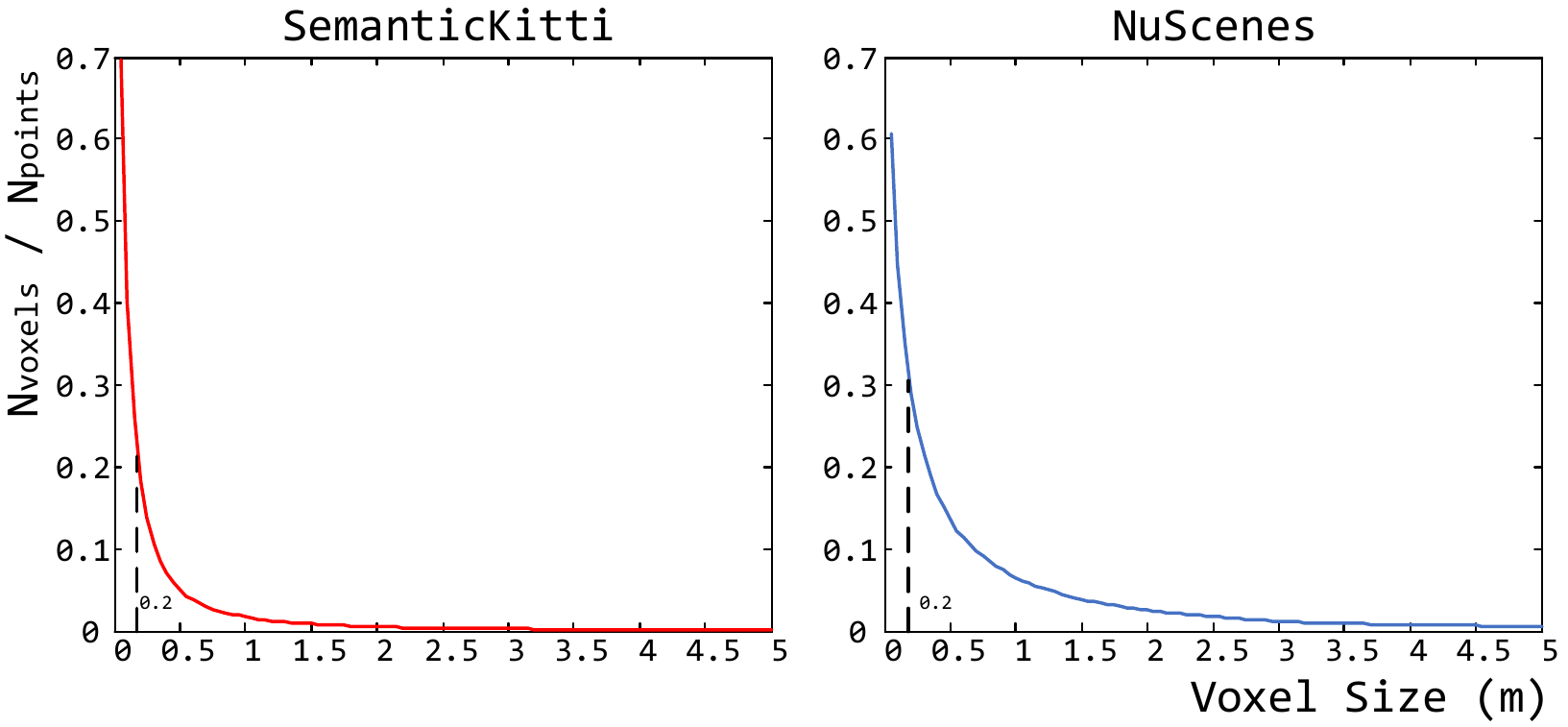}
    \caption{The variation curve of the ratio of voxel number ($N_{voxels}$) to point number ($N_{points}$) against voxel size.}
    \vspace{-5px}
    \label{fig:voxel_number}
\end{figure}

The dual representation learning in DRINet~\cite{ye2021drinet} and SPVCNN~\cite{tang2020searching} includes both pointwise features and voxelwise features. However, the memory consumption increases significantly with more pointwise operations. One key observation is that larger-size voxels contain more points, as shown in Fig.~\ref{fig:voxel_number}, meaning that the number of voxels will greatly decrease with coarser voxels while the number of the points number keeps the same. Another intuition is that voxelwise features are aggregated features within a local neighborhood, which implicitly implies pointwise geometry. Based on these two observations, we can treat each voxel as a special type of ``\textbf{super point}" and apply pointwise operations on voxels namely ``\textbf{voxels as points}'', which can reduce memory cost and improve performance with the pointwise operations. By treating voxels as points, we integrate sparsity and geometry in a novel and efficient way, which is applicable in any multi-representation learning framework. We conduct experiments in sec.\ref{sec:ablation} in several multi-representation approaches to verify this principle. 

\subsection{Sparse Feature Encoder}
The sparse voxel representation makes it easy to apply standard convolution operations to extract local context information. Thus, we utilize sparse convolution layers~\cite{yan2018second, Graham20173D} instead of dense convolution~\cite{liu2019point}. One merit of sparse convolution lies in the sparsity, with which the convolution operation only considers the non-empty voxels. Based on this, we build our sparse feature encoder (SFE) with sparse convolutions to quickly expand the receptive field with less computation cost. We adopt the ResNet BottleNeck~\cite{he2016deep} while replacing the ReLU activation with Leaky ReLU activation~\cite{maas2013rectifier} compared with DRINet~\cite{ye2021drinet} and SPVCNN~\cite{tang2020searching}. In order to keep a high efficiency, all the channel number for sparse convolution is set to $64$.

\subsection{Sparse Geometry Feature Enhancement}
After obtaining the sparse voxelwise features from the sparse feature encoder, we aim to enhance the voxelwise features with more geometric guidance by our Sparse Geometry Feature Enhancement (SGFE).

\textbf{Multi-scale Sparse Projection (MSP).} Inspired by previous works ~\cite{zhao2017pyramid, qiPointNetDeepHierarchical2017a,ye2021drinet} that focus on multi-scale features aggregation, we notice that hierarchical context information helps enhance the capability of feature extraction, especially for point clouds. Multi-scale features bring point cloud learning more geometric enhancement since each voxel scale reflects one specific physical dimension property. 
However, it is not applicable to directly apply Pyramid Pooling Module from PSPNet~\cite{zhao2017pyramid} due to the sparsity of point cloud. DRINet~\cite{ye2021drinet} proposes points max pooling at point level by scattering operation while introducing extra huge memory cost and resulting in information loss. As a result, we propose~\textbf{Multi-scale Sparse Projection} layer to exploit the multi-scale features at a sparse voxel level through voxels as points, with a lower memory cost.

Given the input voxelwise features $F$ and pre-defined scale set $S$, Multi-scale Sparse Projection is described in Algo.~\ref{alg:multi-scale-pooling}. 
For each scale, we project the features into local regions to get projection mean embedding with different local geometric priors. Then features shift that is calculated based on the projection embedding will be treated as the kernel weighted function like KPConv~\cite{thomas2019kpconv} to strengthen input features. This enforces similar features share similar weights for more consistent local geometry. Finally, features at different scales are stacked together.

\begin{figure}
    \centering
    \includegraphics[width=6.5cm]{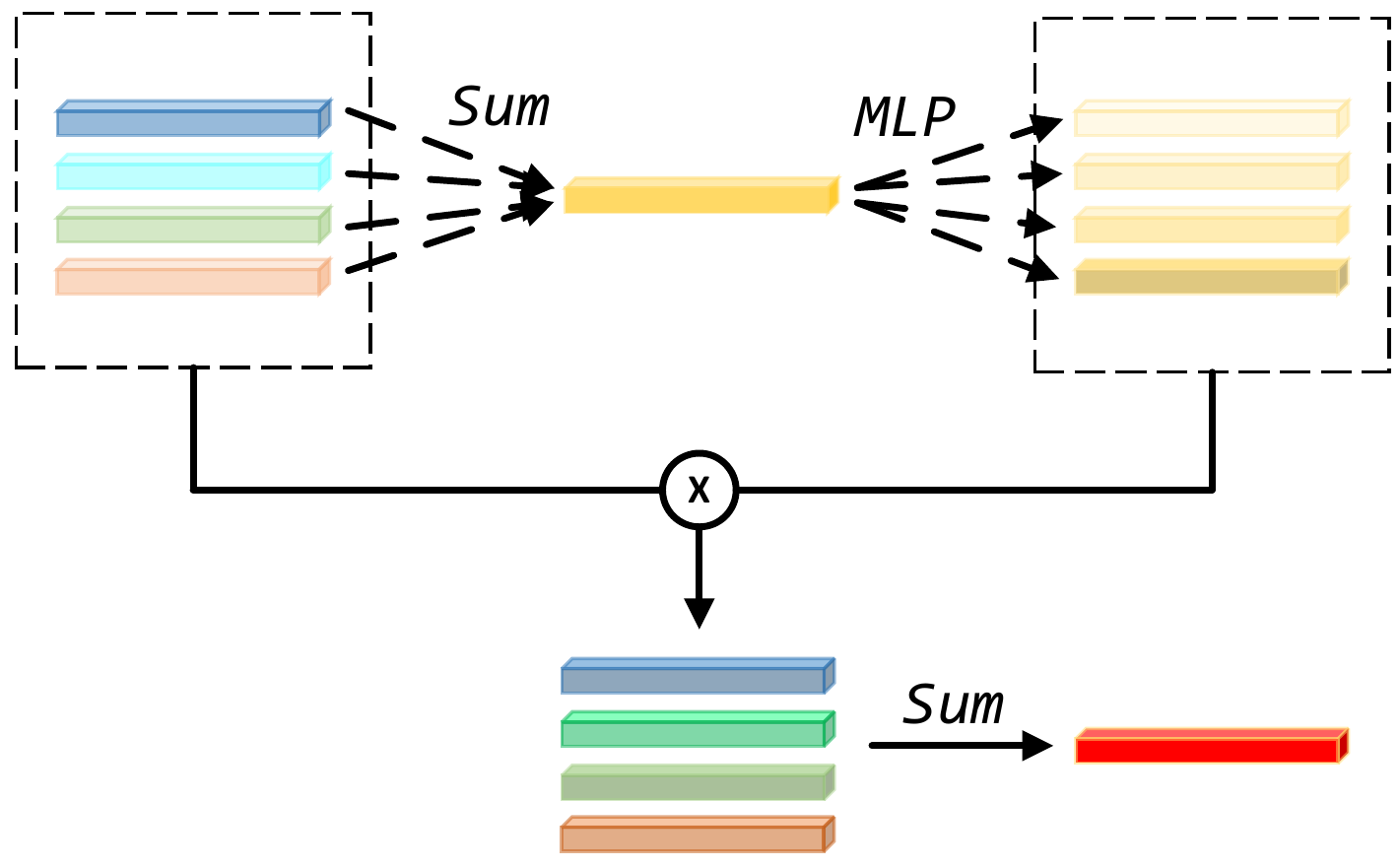}
    \caption{An example demonstrates the Attentive Multi-scale Fusion. $\otimes$ means tensor elementwise multiplication.}
    \label{fig:attentive_selection}
\end{figure}

\begin{algorithm}[tb]
\caption{Multi-scale Sparse Projection}
\label{alg:multi-scale-pooling}
\textbf{Input}: Input sparse voxel features $F$ and pre-defined scale set $S$

\textbf{Output}: Multi-scale features $O$
\begin{algorithmic}[1] 
\STATE $L$ = []
\FOR{each $s\in S$}
\STATE $V^s$ = AvgPooling($F, s$)
\STATE $O^s$ = $F$ $-$ UpsampleNearest($V^s$)
\STATE $O^s$ = $\text{MLP}(O^s) * F$
\STATE $L .\text{append}(\text{MLP}(O^s))$
\ENDFOR
\STATE $O\leftarrow \text{Stack}(L)$
\STATE return $O$


\end{algorithmic}
\end{algorithm}


\textbf{Attentive Multi-scale Fusion (AMF).} After obtaining the multi-scale features from the multi-scale sparse projection layer, a naive way of multi-scale features fusion is to apply tensor concatenation or summation. As such, all the features from different scales share the same weights and are treated equally. We believe that the features from different scales contain different geometric priors and represent a scene hierarchically. From this point of view and motivated by the SENet~\cite{hu2018squeeze} and SKNet~\cite{li2019selective}, a better approach to fusing multi-scale features is to apply re-weighting strategy along scale dimension for each feature channel to re-distribute the importance of each scale. 

As shown in Fig.~\ref{fig:attentive_selection}, we first sum over all the input tensors in the first-stage fusion to collect all the information from different scales. Inspired from current popular attention works~\cite{li2019selective}, we apply a multi-scale MLP layer with sigmoid activation for the results to get the attentive embedding for each scale. Finally, tensor multiplication between attention weight tensor and multi-scale features is applied, followed by the tensor summation among multi-scale features. Compared with AF2S3Net~\cite{cheng20212}, our Attentive Multi-scale Fusion is more lightweight without introducing extra convolution layers and is only operated at the voxel with the voxel-as-point principle for better efficiency.

\begin{algorithm}[tb]
\caption{DRINet++}
\label{alg:overall}
\textbf{Input}: Input sparse voxel features $F$, the number of blocks $B$, and the target scale $s$

\textbf{Output}: Semantic prediction $P$
\begin{algorithmic}[1] 

\STATE $L$ = []
\STATE $Loss$ = 0
\FOR{${i}$ = 1 to $B$}
\STATE $V\leftarrow \text{SFE}(F)$
\STATE $F\leftarrow \text{SGFE}(V)$
\STATE $O^s \leftarrow \text{UpsampleNearest}(F, s)$
\STATE $Loss \mathrel{+}= \text{LossFunc}(V)$
\STATE $L.\text{append}(O^s)$
\ENDFOR
\STATE $L\leftarrow \text{Stack}(L)$
\STATE $P$ = Softmax(MLP($L$))
\STATE return $P$


\end{algorithmic}
\end{algorithm}

\begin{table*}[!t]
    \small
    \centering
    \resizebox{\textwidth}{!}{
\setlength{\tabcolsep}{1.3mm}{            \renewcommand{\arraystretch}{1.2}
    \begin{tabular}{@{}l c c c c c c c c c c c c c c c c c c c c c@{}}
    \hline
    Methods & \rotatebox{90}{road} & \rotatebox{90}{sidewalk} & \rotatebox{90}{parking} & \rotatebox{90}{other ground} & \rotatebox{90}{building} & \rotatebox{90}{car} & \rotatebox{90}{truck} & \rotatebox{90}{bicycle} & \rotatebox{90}{motorcycle} & \rotatebox{90}{other vehicle \,} & \rotatebox{90}{vegetation} & \rotatebox{90}{trunk} & \rotatebox{90}{terrain} & \rotatebox{90}{person} & \rotatebox{90}{bicyclist} & \rotatebox{90}{motorcyclist} & \rotatebox{90}{fence} & \rotatebox{90}{pole} & \rotatebox{90}{traffic sign} & \rotatebox{90}{mIoU} & \rotatebox{90}{speed (ms)}
    \\
    \hline \hline
    PointNet~\cite{qi2017pointnet} & 61.6 & 35.7 &	15.8 & 1.4 & 41.4 &	46.3 & 0.1 & 1.3 & 0.3 & 0.8 & 31.0 & 4.6 & 17.6 & 0.2 & 0.2 & 0.0 & 12.9 & 2.4 & 3.7 & 14.6 & 500
    \\
    PointNet++~\cite{qiPointNetDeepHierarchical2017a} & 72.0 & 41.8 &	18.7 & 5.6 & 62.3 &	53.7 & 0.9 & 1.9 & 0.2 & 0.2 & 46.5 & 13.8 & 30.0 & 0.9 & 1.0 & 0.0 & 16.9 & 6.0 & 8.9 & 20.1 & 5900
    \\
    KPConv~\cite{thomas2019kpconv} & 88.8 & 72.7 &	61.3 & {31.6} & 90.5 &	96.0 & 33.4 & 30.2 & 42.5 & 44.3 & 84.8 &	69.2 & 69.1 & 61.5 & 61.6 & 11.8 & 64.2 & 56.4 & 47.4 & 58.8 & -
    \\
    RandLA~\cite{hu2020randla} & 90.7 & 73.7 & 60.2 & 20.4 & 86.9 & 94.2 & 40.1 & 26.0 & 25.8 & 38.9 & 81.4 & 66.8 & 49.2 & 49.2 & 48.2 & 7.2 & 56.3 & 47.7 & 38.1 & 53.9 & 880
    \\
    SqueezeSegV3~\cite{xu2020squeezesegv3} & 91.7 & 74.8 & 63.4 & 26.4 & 89.0 & 92.5 & 29.6 & 38.7 & 36.5 & 33.0 & 82.0 & 58.7 & 65.4 & 45.6 & 46.2 & 20.1 & 59.4 & 49.6 & 58.9  & 55.9 & 238
    \\
    RangeNet++~\cite{milioto2019rangenet++} & 91.8 & 75.2 & 65.0 & 27.8 & 87.4 & 91.4 & 25.7 & 25.7 & 34.4 & 23.0 & 80.5 & 55.1 & 64.6 & 38.3 & 38.8 & 4.8 & 58.6 & 47.9 & 55.9 & 52.2 & 83.3
    \\
    TangentConv~\cite{tatarchenko2018tangent} &83.9 & 63.9 &33.4 &15.4& 83.4& 90.8& 15.2& 2.7& 16.5& 12.1& 79.5& 49.3& 58.1& 23.0 & 28.4 & 8.1& 49.0& 35.8& 28.5 & 35.9 & 3000
    \\
    SPVCNN~\cite{tang2020searching} & 90.2 & {75.4} & {67.6} & 21.8 & {91.6} & {97.2} & 56.6 & 50.6 & 50.4 & {58.0} & {86.1} & {73.4} & {71.0} & 67.4 & 67.1 & 50.3 & 66.9 & {64.3} & {67.3} & 67.0 & 259
    \\
    PolarNet~\cite{zhang2020polarnet} & 90.8 & 74.4 & 61.7 & 21.7 & 90.0 & 93.8 & 22.9 & 40.3 & 30.1 & 28.5 & 84.0 & 65.5 & 67.8 & 43.2 & 40.2 & 5.6 & 61.3 & 51.8 & 57.5 & 54.3 & {62}
    \\
    DASS~\cite{unal2021improving} & 92.8 & 71.0 & 31.7 & 0.0 & 82.1 & 91.4 & \textbf{66.7} & 25.8 & 31.0 & 43.8 & 83.5 & 56.6 & 69.6 & 47.7 & 70.8 & 0.0 & 39.1 & 45.5 & 35.1 & 51.8 & 90
    \\
    JS3C-Net~\cite{yan2020sparse} & 88.9 &72.1 &61.9& 31.9& 92.5 &95.8 &54.3& 59.3& 52.9& 46.0 &84.5 &69.8 &67.9& 69.5& 65.4& 39.9& 70.8 &60.7 &68.7 & 66.0 & - \\
    DRINet~\cite{ye2021drinet} & 90.7 & 75.2 & 65.0 & 26.2 & 91.5 & 96.9 & 43.3 & {57.0} & {56.0} & 54.5 & 85.2 & 72.6 & 68.8 & {69.4} & {75.1} & {58.9} & {67.3} & 63.5 & 66.0 & {67.5} & {62}
    \\
    Cylinder3D~\cite{zhu2021cylindrical} & 92.0 & 70.0 & 65.0 & 32.3 & 90.7 & 97.1 & 50.8 & 67.6 & {63.8} & 58.5 & 85.6 & 72.5 & 69.8 & {73.7} & {69.2} & {48.0} & {66.5} & 62.4 & 66.2 & {68.9} & {131}
    \\
    AF2S3Net~\cite{cheng20212} & 92.0 & 76.2 & 66.8 & \textbf{45.8} & 92.5 & 94.3 & 40.2 & 63.0 & \textbf{81.4} & 40.0 & 78.6 & 68.0 & 63.1 & 76.4 & 81.7 & \textbf{77.7} & 69.6 & 64.0 & \textbf{73.3} & \textbf{70.8} & -
    \\
    RPVNet~\cite{xu2021rpvnet} & \textbf{93.4} & \textbf{80.7} & \textbf{70.3} & 33.3 & \textbf{93.5} & \textbf{97.6} & 44.2 & \textbf{68.4} & 68.7 & \textbf{61.1} & 86.5 & \textbf{75.1} & 71.7 & 75.9 & 74.4& 43.4 & \textbf{72.1} & 64.8 & 61.4 & 70.3 & 168
    \\ \hline
    Ours & 89.8 & 74.6  & 66.2 & 30.1 & 92.3 & 96.9& 59.3 &	{65.8} & 58.0 & 61.0 & \textbf{87.3} & 73.0& \textbf{72.5} & \textbf{80.4} &	\textbf{82.7} &	46.3 &	69.6 & \textbf{66.1} & 71.6 & 70.7$\ddagger$ & \textbf{59}
    \\
    \hline
    \end{tabular}}}
    \vspace{-1mm}
    \caption{The per-class mIoU results on the SemanticKITTI test set. $\ddagger$ donates the second best results. All results are obtained from the literature or leaderboard, including standard Test-Time Augmentation.}
    \label{semantic_kitti}
\end{table*}

\subsection{Deep Sparse Supervision} 
\label{sec:dss}
Dense supervision on each pixel/voxel is a popular way in both 2D and 3D semantic segmentation tasks. Previous methods ~\cite{zhang2020polarnet,zhu2021cylindrical} generate dense feature maps with dense supervision. Although these works have considered the sparsity with their network architectures, they ignore this property when designing the loss, one of the key differences between 2D and 3D data. In fact, dense supervision with fine-grained feature maps brings a significant overload on memory usage. For example, when using a grid size of $0.2m$, the memory consumption (about $500$ Mb) for a single dense feature map with 20 classes could be problematic. Based on these observations and inspired by~\cite{loquercio2020learning}, we propose a novel~\textbf{Deep Sparse Supervision (DSS)} to deal with supervision in a deep sparse style. 

Since we stack multiple blocks of SFE to generate sparse voxelwise features at different scales, we apply sparse supervision to the output voxelwise features stage by stage as a deep auxiliary loss. We also apply sparse supervision for the main final prediction branch. The auxiliary loss helps optimize the training, while the main branch loss accounts for the most gradients.

In the testing stage, all the auxiliary branches are disabled to keep the runtime efficiency. This training strategy has been proven its effectiveness in image-based segmentation~\cite{long2015fully}. We consider the sparsity of point cloud and apply it in a sparse manner to save memory consumption.

\subsection{Final Prediction} For the final semantic prediction, we fuse the multi-stage features from the output of each Sparse Geometry Feature Enhancement Layer by nearest upsampling to the most fine-grained scale of the voxel. To obtain pointwise results, we also apply the nearest interpolation strategy, with which each point is attached with the semantic features from the voxel that it lies in. The whole algorithm for our framework is illustrated in Algo.~\ref{alg:overall}.

\begin{table}[!t]
\vspace{-5px}
    \small
    \centering
    \begin{tabular}{l  c  c }
    \hline
    Method &  mIoU & Acc
    \\
    \hline \hline
    Cylinder3D~\cite{zhu2021cylindrical} & 52.5 & {91.0} \\
    TemporalLidarSeg~\cite{duerr2020lidar} & 47.0 & - \\
    KPConv~\cite{thomas2019kpconv} & 51.2 & 89.3 \\
    LatticeNet~\cite{rosu2019latticenet} & 45.2 & 89.3 \\
    AF2S3Net~\cite{cheng20212} & 56.9 & 88.1 \\
    \hline
    Ours & \textbf{61.3} & \textbf{92.4} \\
    \hline
    \end{tabular}
    \vspace{-5px}
\caption{Comparison to the state-of-the-art methods on the test set of SemanticKITTI multiple scans challenge.}
\label{tab:multiscan}
\end{table}

\section{Experiments}
\subsection{Experimental Setup}
We conduct experiments on two large-scale outdoor datasets, SemanticKITTI~\cite{behley2019semantickitti} and Nuscenes~\cite{caesar2020nuscenes}, to evaluate our proposed approach.

\textbf{SemanticKITTI}~\cite{behley2019semantickitti}\textbf{.} The SemanticKITTI dataset is generated from the KITTI dataset~\cite{geiger2013vision} and contains 22 sequences that involve common scenes for autonomous driving. Each scan in the dataset has more than 100K points on average with pointwise annotation labels for 20 classes. According to the official settings, sequences from 00 to 10 except 08 are the training split, sequence 08 is the validation split, and the rest sequences from 11 to 21 are the test split.

\textbf{Nuscenes}~\cite{caesar2020nuscenes}\textbf{.} The Nuscenes dataset has a total of 40,000 scans collected by a 32-beam LiDAR. Compared with SemanticKITTI~\cite{behley2019semantickitti}, it contains fewer points and annotations classes (16 classes).

Both datasets use \textbf{mIoU} as one evaluation metric, which is popular in point cloud semantic segmentation. In addition, SemanticKITTI provides the average accuracy metric \textbf{Acc}, and Nuscenes uses \textbf{fwIoU} as additional criteria, which is the weighted sum of IoU for each class based on point-level frequencies.

\textbf{Network Details.} We use the same setting for both datasets. We discretize a point cloud with a voxel scale of $0.2m$ along $xyz$ dimensions to generate the initial sparse voxel features. Our DRINet++ has four blocks of Sparse Feature Encoder and Sparse Geometry Feature Enhancement. In the ablation study, we also evaluate the effect of block number choice between performance and efficiency. As for the multi-scale sparse projection layer, we adopt kernel sizes and strides both with $[2, 4, 8, 16]$, which could cover the coarse and the fine pooling regions. Similar to previous works, we apply random flipping, random point dropout, random scale, and global rotation in the training stage. Besides that, motivated by the improvement achieved by Pseudo Label~\cite{lee2013pseudo} in semi-supervised learning, we apply this technique for our data augmentation. First, we train our model on the training set, and then we predict the pointwise label for the validation set as the Pseudo label. After merging the training set and validation set with pseudo labels, we train our model with this mixture dataset to improve the performance. 
For the loss design, we follow Cylinder3D~\cite{zhu2021cylindrical} to combine the Lovasz loss~\cite{berman2018lovasz} and cross-entropy loss as supervision. Adam optimizer~\cite{kingma2014adam} is employed with an initial learning rate of $2{e^{ - 3}}$ at batch size 4 for $50$ epochs. Learning rate decays in a ratio of $0.1$ for every $15$ epochs.

\begin{table}[!t]
\vspace{-5px}
   \begin{center}
      \small
      \centering
      \setlength{\tabcolsep}{4pt}
      \begin{tabular}{@{}lccccc@{}}
      \hline 
      Method & Param (M) & Macs (G) & Speed & Memory & mIoU \\
      \hline \hline
      SPVCNN & 12.5 & 73.8 & 120 ms & 2.4 Gb & 67.0 
      \\
      DRINet & 3.5 & 14.58 & 62$^*$ ms & 2.1$^*$ Gb & 67.5
      \\
      Cylind3D & 53.3 & 64.3 & 131 ms & 3.0 Gb & 68.9
      \\
      RPVNet & 24.8 & 119.5 & 168$^*$ ms & 2.7$^*$ Gb & 70.3
      \\ \hline 
      Ours & \textbf{2.2} & \textbf{12.1} & \textbf{59 ms} & \textbf{1.4 Gb} & \textbf{70.7}
      \\
      \hline
      
      \end{tabular}
   \end{center}
   \vspace{-1mm}
   \caption{Quantitative results of model complexity with the performance from the leaderboard of SemanticKITTI on the test set. Statistics of the number of parameters and Macs are from the corresponding papers. The speed of each method is evaluated on a single Nvidia RTX 2080Ti GPU. $*$ donates the statistics are from our reproduction.}
\label{tab:efficiency}
\vspace{-10px}
\end{table}

\begin{table*}[!t]
\vspace{-10px}
    \small
    \centering
    \resizebox{\textwidth}{!}
{
\renewcommand{\arraystretch}{1.2}
    \begin{tabular}{@{}l c c c c c c c c c c c c c c c c c c@{}}
    \hline
    Method & \rotatebox{90}{barrier} & \rotatebox{90}{bicycle} & \rotatebox{90}{bus} & \rotatebox{90}{car} & \rotatebox{90}{construction} & \rotatebox{90}{motorcycle} & \rotatebox{90}{pedestrian} & \rotatebox{90}{traffic cone} & \rotatebox{90}{trailer} & \rotatebox{90}{truck} & \rotatebox{90}{driveable} & \rotatebox{90}{other\_flat} & \rotatebox{90}{sidewalk} & \rotatebox{90}{terrain} & \rotatebox{90}{manmade} & \rotatebox{90}{vegetation} & \rotatebox{90}{FW mIoU} & \rotatebox{90}{mIoU} 
    \\
    \hline \hline
    AF2S3Net~\cite{cheng20212} & 78.9& \textbf{52.2}& 89.9& 84.2& \textbf{77.4}& 74.3& 77.3& 72.0& 83.9& 73.8& 97.1& 66.5& 77.5& 74.0& 87.7& 86.8& 88.5& 78.3 \\
    Cylinder3D~\cite{zhu2021cylindrical} & 82.8& 29.8& 84.3& 89.4& 63.0& 79.3& 77.2& \textbf{73.4}& \textbf{84.6}& 69.1& \textbf{97.7}& 70.2& 80.3& 75.5& 90.4& 87.6& 89.9& 77.2\\
    SPVCNN~\cite{tang2020searching} & 80.0& 30.0& \textbf{91.9}& 90.8& 64.7& 79.0& 75.6& 70.9& 81.0& 74.6& 97.4& 69.2& 80.0& 76.1& 89.3& 87.1& 89.7& 77.4 \\ 
    PolarNet~\cite{zhang2020polarnet} & 72.2& 16.8& 77.0& 86.5& 51.1& 69.7& 64.8& 54.1& 69.7& 63.5& 96.6& 67.1& 77.7& 72.1& 87.1& 84.5& 87.4& 69.4 \\ \hline
    Ours & \textbf{85.5}& 43.2& 90.5& \textbf{92.1}& 64.7& \textbf{86.0}& \textbf{83.0}& 73.3& 83.9& \textbf{75.8}& 97.0& \textbf{71.0}& \textbf{81.0}& \textbf{77.7}& \textbf{91.6}& \textbf{90.2} & \textbf{91.0} & \textbf{80.4} \\
    \hline
     \end{tabular}
    }
    \vspace{-1mm}
    \caption{The per-class mIoU results on the Nuscenes test set. Note that our DRINet++ shows result without \textit{Pseudo label}.}
    \label{tab:nuscenes}
\end{table*}

\begin{figure*}[t]
    \centering
    \includegraphics[width=1.0\textwidth]{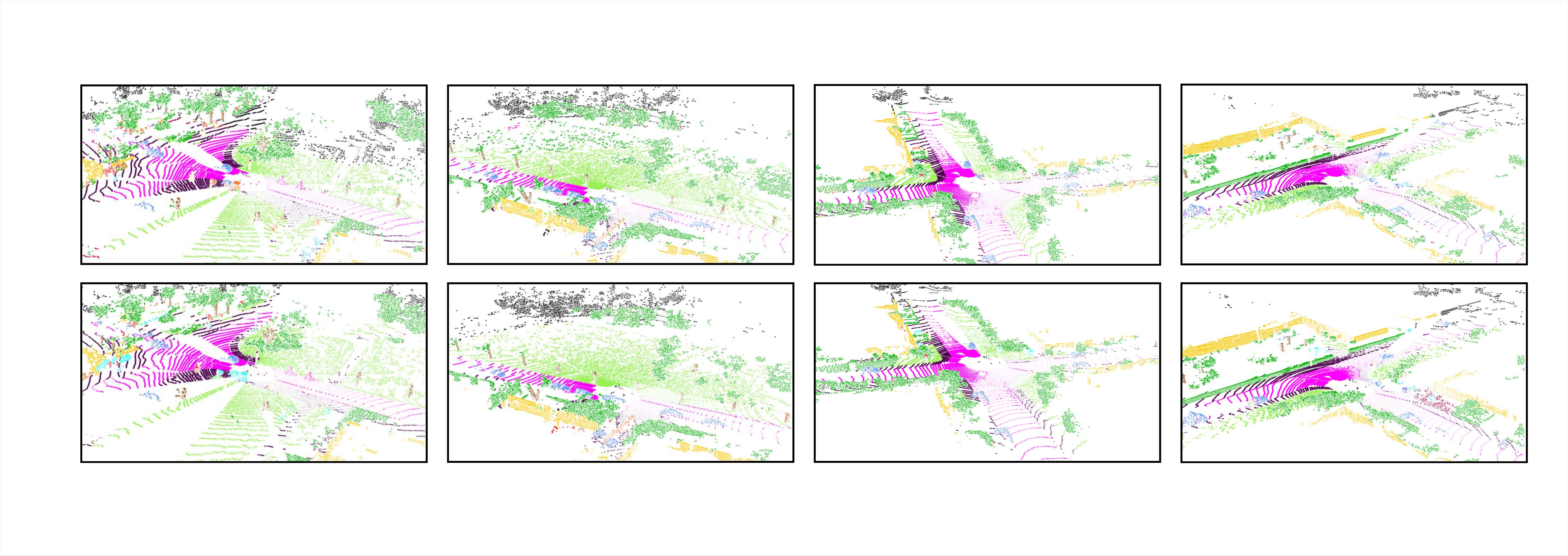}
    \caption{The results on the SemanticKITTI valid set. The top row is the ground truth, and the bottom row is the predictions by DRINet++.}
    \label{fig:semantic_results}
\end{figure*}

\subsection{Results on SemanticKITTI}
We provide the detailed per-class quantitative results of our DRINet++ as well as other state-of-the-art methods in Tab.~\ref{semantic_kitti}. Compared with previous methods, DRINet++ achieves state-of-the-art performance while maintaining real-time inference efficiency. Although DRINet++ fails to achieve the best result for every class, it achieves balanced results among all the classes. Even compared to multiple representation fusion approaches~\cite{ye2021drinet, tang2020searching, xu2021rpvnet}, our method still surpasses by a considerable margin.

Furthermore, we also provide quantitative analysis for the model complexity and latency for some state-of-the-art methods and our DRINet++ in Tab.~\ref{tab:efficiency} to illustrate the high performance-time ratio of our approach. Compared with previous methods, we achieve the best mIoU result with the least computation cost, demonstrating the efficiency and effectiveness of our approach. Some qualitative results on the SemanticKITTI validation set are shown in Fig.~\ref{fig:semantic_results}.

\textbf{Multiple Scans.} Meanwhile, we also conduct experiments on SemanticKITTI multiple scans challenge to verify the effectiveness of our DRINet++. In this task, we directly stack multiple aligned scans as input without any temporal fusion algorithm, then generate the pointwise output prediction according to Algo.~\ref{alg:overall}. We do not apply any post-processing for refinement. As shown in Tab.~\ref{tab:multiscan}, our DRINet++ shows a significant improvement compared with previous methods in terms of both metrics. Compared with AF2S3Net~\cite{cheng20212}, which is a voxel-based approach, the proposed method brings a very competitive gain (about $4.4\%$). 

\begin{table}
   \begin{center}
      \small
      \centering
      \begin{tabular}{ccc}
      \hline
      Number of blocks & mIoU (\%) & Speed (ms) \\
      \hline \hline
      1 & 52.0 & 26 \\
      \hline
      2 & 62.3 & 35 \\ 
      \hline
      3 & 68.4 & 47 \\
      \hline
      4 & 70.8 & 59 \\
      \hline
      5 & 71.2 & 70 \\
      \hline
      \end{tabular}
   \end{center}

\caption{Controlled experiments for the number of blocks on the SemanticKITTI validation set.}
\vspace{-10px}
\label{tab:block_num}
\end{table}

\subsection{Results on Nuscenes}
To verify the generalization ability of our approach, we also report the results on the test set of the Nuscenes-lidarSeg~\cite{caesar2020nuscenes} task. As shown in Tab.~\ref{tab:nuscenes}, our DRINet++ achieves highly competitive results, with $10$ out of $16$ categories surpassing all other approaches. DRINet++ has a noticeable improvement for classes with small sizes, such as pedestrians and motorcycles. This demonstrates the effectiveness of our sparse geometry feature enhancement, which aims to capture and integrate multi-scale context information through hierarchical feature learning and attentive multi-scale fusion.  

\begin{table}
   \begin{center}
      \small
      \centering
      \begin{tabular}{c|c c|c|c|c}
      \hline
      \multirow{2}* {SFE} & \multicolumn{2}{c|}{SGFE} & \multirow{2}* {DSS} & \multirow{2}* {Pseudo label} & \multirow{2}* {mIoU (\%)} \\
        \cline{2-3} & MSP & AMF & & & \\
      \hline \hline
      \checkmark & & & & & 64.5\\
      \hline
      \checkmark &\checkmark & & & & 65.8\\
      \hline
      \checkmark & \checkmark&\checkmark & & & 67.9\\
      \hline
      \checkmark & \checkmark&\checkmark & \checkmark& & 69.4\\
      
      \hline
      \checkmark & \checkmark& \checkmark& \checkmark& \checkmark& 70.8 \\
      \hline
      \end{tabular}
   \end{center}

\caption{Ablation study on the SemanticKITTI validation set. MSP refers to Multi-scale Sparse Projection. AMF refers to Attentive Multi-scale Fusion. DSS refers to Deep Sparse Supervision.}
\vspace{-10px}
\label{tab:ablation_study}
\end{table}

\subsection{Ablation Study}
\label{sec:ablation}
We conduct ablation studies to analyze the effectiveness of each proposed component on the validation set of SemanticKITTI. 

\textbf{Component Study.} Our baseline model, shown in Tab.~\ref{tab:ablation_study}, which only uses SFE, has achieved decent performance. This shows that the voxel representation with sparse convolution is highly capable of feature extraction and context information learning in outdoor point cloud semantic segmentation. The MSP layer brings $1.3\%$ improvement with its ability to capture hierarchical geometry information. Furthermore, we apply the AMF to re-distribute the importance of each scale for each channel. This strategy enables the network the focus on the more significant scale and has a performance gain of about $2.1\%$. Next, adding the deep sparse supervision in the training stage improves the mIoU to $69.4\%$, an increase of $1.5\%$. More importantly, this supervision will be disabled when inference, bringing no extra computation cost for deployment. Finally, the data augmentation with pseudo labels shows a promising improvement by leveraging the potential of semi-supervised learning, giving a performance boost of about $1.4\%$. 

\begin{table}[!t]
   \begin{center}
      \small
      \centering
      \begin{tabular}{cccc}
      \hline
       & VAP & Memory & mIoU (\%)\\
      \hline \hline
      \multirow{2}{*}{DRINet++} & $\times$ & 2.0 Gb & 71.0 \\
       & \checkmark& 1.4 Gb & 70.8 \\
      \hline
      \hline
      \multirow{2}{*}{DRINet~\cite{ye2021drinet}} &$\times$ & 2.1 Gb & 67.3 \\
       & \checkmark& 1.3 Gb & 67.2 \\
      \hline
      \hline
      \multirow{2}{*}{SPVCNN~\cite{tang2020searching}} &$\times$ & 2.4 Gb & 64.7 \\
       & \checkmark& 1.9 Gb & 64.6 \\
      \hline
      
      \end{tabular}
   \end{center}
\vspace{-10px}
\caption{Ablation study for the voxel-as-point (VAP) design on the SemanticKITTI validation set.}

\label{tab:vap}
\end{table}
\textbf{Voxel-as-point Analysis.} To verify the voxels-as-points (VAP) principle, we compare different models under the VAP and non-VAP settings. As shown in Tab.~\ref{tab:vap}, the VAP design saves about $30\%$ memory usage while only leading to a decrease of about $0.2\%$ mIoU, indicating that the VAP can be a valuable and general principle for the multi-representation framework. By VAP, models can be improved and be lighter without real point representation.

\textbf{Number of Blocks.} 
We conduct an experiment with various numbers of blocks. As shown in Tab.~\ref{tab:block_num}, the mIoU increases from $52.0\%$ to $71.2\%$ when the number of blocks increases to $5$ and the runtime nearly doubles. Furthermore, adding one more block when the block number is $4$ brings extra $11ms$ latency and improves mIoU by about $0.4\%$. Therefore, we choose block number as $4$ in our settings.

\begin{table}[!t]
   \begin{center}
      \small
      \centering
      \begin{tabular}{l|c}
      \hline
      Method & mIoU(\%) \\
      \hline \hline
      AMF & \textbf{70.8} \\ 
      \hline
      Tensor summation & 68.8 \\
      \hline
      Tensor concatenation & 69.6 \\
      \hline
      \end{tabular}
   \end{center}
\vspace{-10px}
\caption{Ablation study for fusion strategy. 
\vspace{-10px}
}

\label{tab:fusion}
\end{table}

\textbf{Scale Feature Fusion Strategy.} We analyze the effectiveness of the proposed AMF by comparing it with different fusion strategies.
We use tensor concatenation and summation, which are commonly used in feature fusion. For this experiment, we enable all other proposed modules for a fair comparison. As shown in Tab.~\ref{tab:fusion}, the AMF layer serves as a better approach for fusing the multi-scale features, which leads to a $2.0\%$ increase compared with the tensor summation.  
Tab.~\ref{tab:supervision_on_models} also shows that other models that utilize multiple representations could benefit from the AMF strategy. Compared with DRINet~\cite{ye2021drinet}, original SPVCNN~\cite{tang2020searching} does not have hierarchical learning in the pointwise branches, and then its performance will improve when AMF is enabled.

\begin{table}[!t]
   \begin{center}
      \small
      \centering
      \begin{tabular}{c|c|c}
      \hline
      Method & GPU Memory (Mb) & mIoU (\%)\\
      \hline \hline
      Sparse & 6 & 67.9 \\
      \hline
      Dense & 552 & 68.0 \\
      \hline
      
      \end{tabular}
   \end{center}
\vspace{-10px}
\caption{Ablation study for supervision ways. }

\label{tab:supervision}
\end{table}
\textbf{Sparse or Dense Supervision.} Deep Sparse supervision is another characteristic of our DRINet++. In this experiment, we compare the sparse and dense supervision in terms of memory cost and performance. We remove the deep auxiliary loss branch for simplicity, with only the main supervision left. The memory consumption for both sparse and dense supervision only includes prediction tensor and label tensor without the consumption used by gradient tensors. As shown in Tab.~\ref{tab:supervision}, the memory footprint in sparse supervision only accounts for about one percent of that in dense supervision, and the results are close between the methods. By incorporating sparse supervision, we can use a larger batch size for training, leading to efficient training. 

\begin{table}[!t]
   \begin{center}
      \small
      \centering
      \begin{tabular}{c|c|c|c}
      \hline
      Method & Original (\%) & DSS (\%) & AMF (\%) \\
      \hline \hline
      SPVCNN~\cite{tang2020searching} & 64.7 & 66.1 (\textbf{+1.4})& 66.7 (\textbf{+2.0}) \\
      \hline
      DRINet~\cite{ye2021drinet} & 67.3 & 68.1 (\textbf{+0.8}) & 67.9 (\textbf{+0.6}) \\
      \hline
      Cylinder3D~\cite{zhu2021cylindrical} &66.5 & 67.8 (\textbf{+1.3})& - \\
      \hline
      \end{tabular}
   \end{center}
\vspace{-10px}
\caption{Ablation study for DSS on the SemanticKITTI~\cite{behley2019semantickitti} valid set with different models.
The statistics are from our reproduction.}

\label{tab:supervision_on_models}
\end{table}

\textbf{Deep Sparse Supervision.} Since our Deep Sparse Supervision is a universal component in the point cloud semantic segmentation task, and we incorporate this strategy with other popular models to verify its effectiveness. As shown in Tab.~\ref{tab:supervision_on_models}, Deep Sparse Supervision could help the performance of popular models without any extra computation cost for inference.

\section{Conclusion}
In this paper, we propose DRINet++, an efficient network architecture for point cloud segmentation via the voxel-as-point principle. 
DRINet++ consists of Sparse Feature Encoder (SFE) and Sparse Geometry Feature Enhancement (SGFE) to fully utilize the sparsity and geometry in a single sparse voxel representation to maintain performance and efficiency. 
SFE extracts local context information, while SGFE enhances the geometry with multi-scale sparse projection and attentive multi-scale fusion. Moreover, deep sparse supervision is applied to accelerate convergence with a lower memory cost. The experiments on large-scale outdoor datasets show that our approach achieves state-of-the-art performance with impressive runtime efficiency.

{\small
\bibliographystyle{ieee_fullname}
\bibliography{egbib}
}

\end{document}



\maketitle
\section{Contents}
In this supplementary material, we provide:
\begin{itemize}
\item More Ablation Study in Sec.~\ref{sec:ablation}.
 \item Detailed test results on the SemanticKITTI multiple scans challenge in Sec.~\ref{sec:detailed_results}. 
 \item Qualitative results of our approach in Sec.~\ref{sec:analyzes}.
 \item Sample codes in Sec.~\ref{sec:sample_codes}.
 \item Accompanying files including the model implementation with its environment requirements and a video file named \textbf{2504.mp4} to show some visual results on the SemanticKITTI test set.
\end{itemize}

\section{Ablation Study}
\label{sec:ablation}
\subsection{Loss Varietas}
Loss choice is also a critical part of the semantic segmentation task, especially for the dataset with the imbalanced distribution. We also conduct experiments for the loss design in the SemanticKITTI~\cite{behley2019semantickitti} validation set.  

\begin{table}[htbp]
    \small
    \centering
    \def\arraystretch{1.1}
    \begin{tabular}{c|c|c|c}
    \hline
     &  CE Loss & Lovasz Loss & CE $+$ Lovasz Loss
    \\
    \hline
    mIoU(\%) & 68.9  & 70.0 & \textbf{70.8}
    \\
    \hline
    \end{tabular}
\vspace{1px}
    \caption{Comparison with different losses. CE is cross entropy loss.}
    \label{tab:loss}
    \vspace{-10px}
\end{table}

As shown in Tab.~\ref{tab:loss}, we can find that Lovasz Loss~\cite{berman2018lovasz} improves the results by nearly $1.1\%$ compared with cross entropy loss. Moreover, by combining these two losses, there is an obvious performance increase about $0.8\%$.

\subsection{Loss Weights}
\begin{table}
   \begin{center}
      \small
      \centering
      \begin{tabular}{cc}
      \hline
      $\alpha$ & mIoU (\%)\\
      \hline
      0.25 & 70.0 \\
      \hline
      0.5 & 70.5 \\ 
      \hline
      1.0 & \textbf{70.8} \\
      \hline
      1.25 & 70.6 \\
      \hline
      1.5 & 70.6 \\
      \hline
      \end{tabular}
   \end{center}

\vspace{1px}
\caption{Ablation study on SemanticKITTI validation set for loss weight.}
\label{tab:alpha}
\vspace{-10px}
\end{table}
Since we apply Deep Sparse Supervision in the training phase, we need to consider the weight of the auxiliary loss. We donate the $\alpha$ as the factor of the auxiliary loss. According to the experimental results from Tab.~\ref{tab:alpha}, we finally set $\alpha$ to $1.0$, which implies that the auxiliary loss should be treated equally as the main loss.

\section{Detailed Results}
\label{sec:detailed_results}
For the SemanticKITTI~\cite{behley2019semantickitti} multi-scan challenge, we achieve significant improvement compared with previous methods. As shown in Tab.~\ref{tab:metricsKitti}, the performance of our GASN surpasses the existing approaches in nearly all categories, which demonstrates the effectiveness and generalization capability of our method. To this end, our GASN can serve as an efficient and strong backbone for the large-scale point cloud semantic segmentation task.  
\begin{table*}%
    \centering%
    \addtolength{\tabcolsep}{-2.5pt}%
    \def\arraystretch{1.25}%
    \resizebox{\textwidth}{!}
{
\begin{tabular}{ l | c  c c c c c c c c c c c c c c c c c c c c c c c c c}
  \hline
Approach & \rotatebox{90}{road} & \rotatebox{90}{sidewalk} & \rotatebox{90}{parking} & \rotatebox{90}{other ground} & \rotatebox{90}{building} & \rotatebox{90}{car} &\rotatebox{90}{car$^*$} & \rotatebox{90}{truck} &\rotatebox{90}{truck$^*$}& \rotatebox{90}{bicycle} & \rotatebox{90}{motorcycle} & \rotatebox{90}{other vehicle \,} &\rotatebox{90}{other vehicle $^*$\,}&  \rotatebox{90}{vegetation} & \rotatebox{90}{trunk} & \rotatebox{90}{terrain} & \rotatebox{90}{person} & \rotatebox{90}{person$^*$} & \rotatebox{90}{bicyclist} &\rotatebox{90}{bicyclist$^*$}& \rotatebox{90}{motorcyclist} &\rotatebox{90}{motorcyclist$^*$} & \rotatebox{90}{fence} & \rotatebox{90}{pole} & \rotatebox{90}{traffic sign} & \rotatebox{90}{mIoU} \\
  \hline \hline
  Cylinder3D & 90.7& 74.5& 65.0& 32.3& 92.6& 94.6& 74.9& 41.3& 0.0& \textbf{67.6}& 63.8& 38.8& 0.1& 85.8& 72.0& 68.9& 12.5& 65.7& 1.7& 68.3& 0.2& 11.9& 66.0& 63.1& 61.4& 52.5 \\
  TemporalLidarSeg & 91.8& 75.8& 59.6& 23.2& 89.8& 92.1& 68.2& 39.2& 2.1& 47.7& 40.9& 35.0& 12.4& 82.3& 62.5& 64.7& 14.4& 40.4& 0.0& 42.8& 0.0& 12.9& 63.8& 52.6& 60.4& 47.0\\
  KPConv & 86.5& 70.5& 58.4& 26.7& 90.8& 93.7& 69.4& 42.5& 5.8& 44.9& 47.2& 38.6& 4.7& 84.6& 70.3& 66.0& 21.6& 67.5& 0.0& 67.4& 0.0& 47.2& 64.5& 57.0& 53.9& 51.2 \\
  AF2S3Net  & 91.3& 72.5& 68.8& \textbf{53.5}& 87.9& 91.8& 65.3& 15.7& 5.6& 65.4& \textbf{86.8}& 27.5& 3.9& 75.1& 64.6& 57.4& 16.4& 67.6& \textbf{15.1}& 66.4& \textbf{67.1}& 59.6& 63.2& 62.6& 71.0& 56.9 \\
  \hline 
   Ours & \textbf{92.3}& \textbf{79.1}& \textbf{69.6}& 30.9& \textbf{93.7}& \textbf{97.2}& \textbf{85.4}& \textbf{46.8}& \textbf{15.9}& 63.6& 53.3& \textbf{64.6}& \textbf{26.3}& \textbf{86.8}& \textbf{75.8}& \textbf{71.2}& \textbf{30.5}& \textbf{84.8}& 0.0& \textbf{73.1}& 0.0& \textbf{76.9}& \textbf{72.8}& \textbf{68.0}& \textbf{73.5}& \textbf{61.3} \\
  \hline  
\end{tabular}
}
\caption{Comparison to state-of-the-art methods including Cylinder3D~\cite{zhu2021cylindrical}, TemporalLidarSeg~\cite{duerr2020lidar}, KPConv~\cite{thomas2019kpconv} and AF2S3Net~\cite{cheng20212} on the SemanticKITTI~\cite{behley2019semantickitti} multiple scans challenge. $*$ donates the moving category.}
\label{tab:metricsKitti}
\end{table*}%

\begin{figure*}[t]
    \centering
    \includegraphics[width=1.0\textwidth]{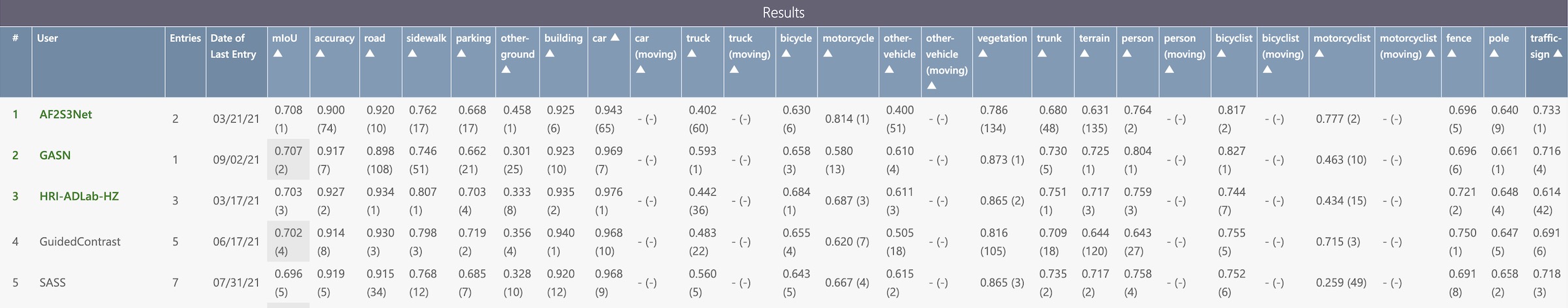}
    \caption{The ranking of SemanticKITTI single scan challenge.}
    \label{fig:single_scan}
\end{figure*}

\begin{figure*}[t]
    \centering
    \includegraphics[width=1.0\textwidth]{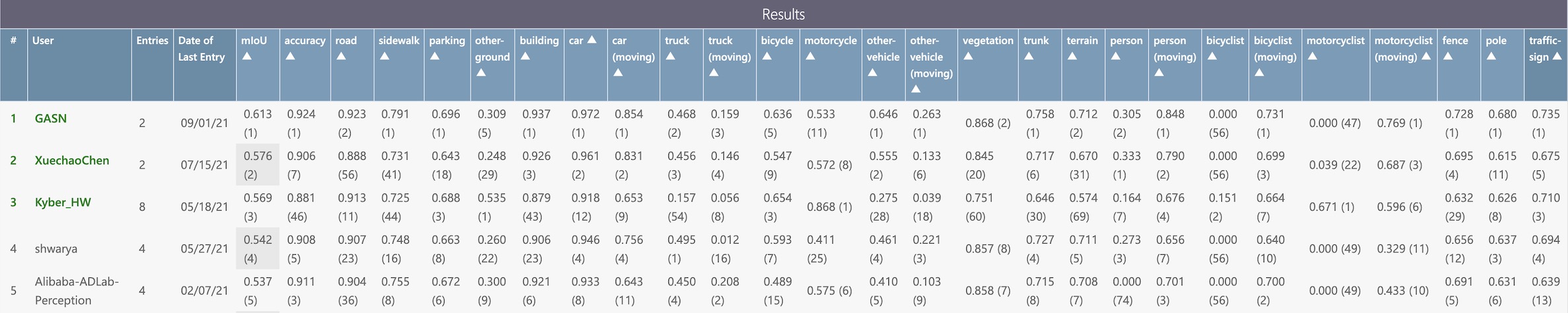}
    \caption{The ranking of SemanticKITTI multiple scan challenge.}
    \label{fig:single_multiscan}
\end{figure*}

\begin{figure*}[t]
    \centering
    \includegraphics[width=1.0\textwidth]{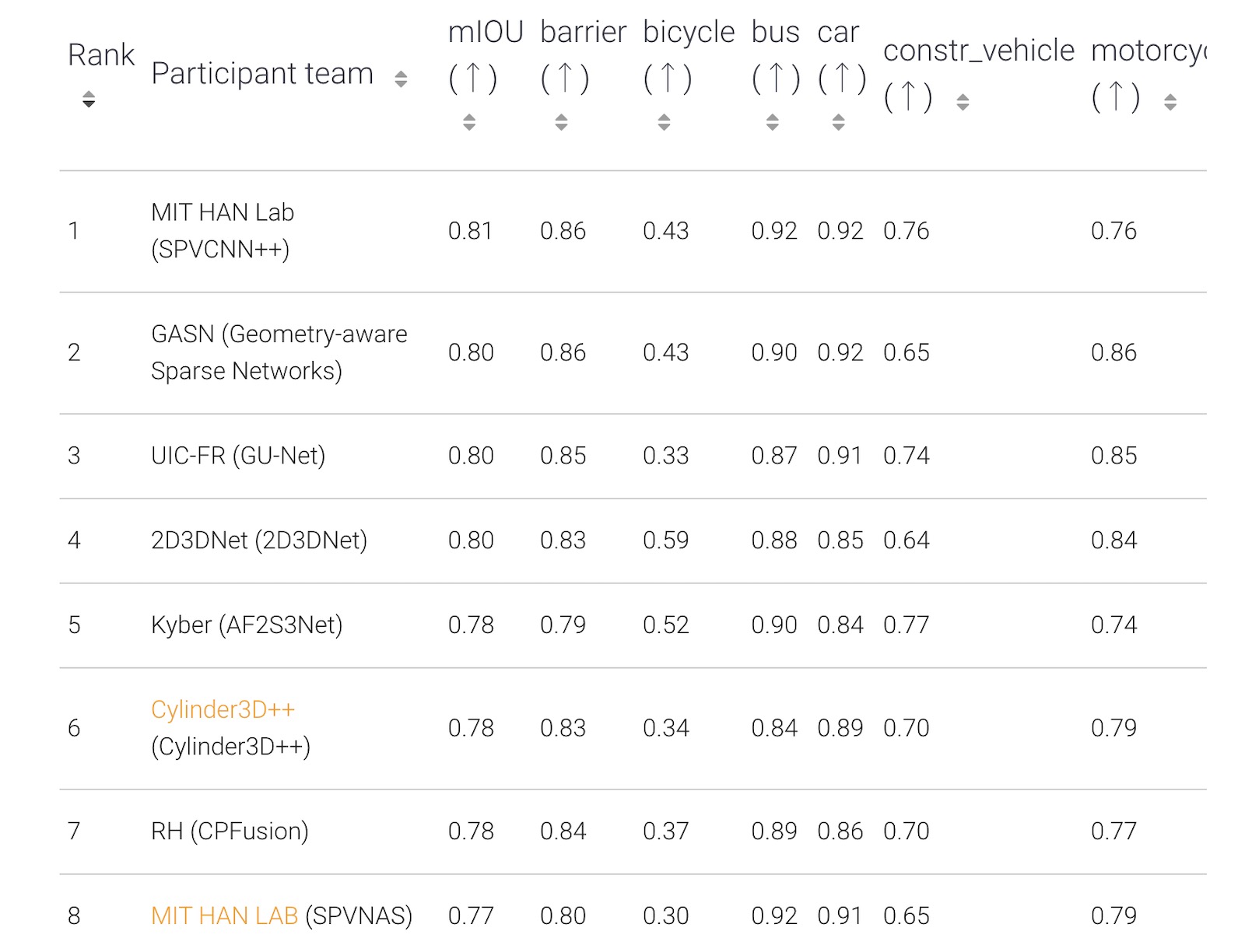}
    \caption{The ranking of Nuscenes LidarSeg challenge.}
    \label{fig:nuscenes_seg}
\end{figure*}

Moreover, we provide some snapshots of our method in different public open challenges in Fig.~\ref{fig:single_scan}, Fig.~\ref{fig:single_multiscan} and Fig.~\ref{fig:nuscenes_seg} on September 11th 2021.
\section{Qualitative Analysis}
\label{sec:analyzes}
In this section, we provide some qualitative results of our approach in the test set of both Nuscenes~\cite{caesar2020nuscenes} and SemanticKITTI dataset~\cite{behley2019semantickitti}.
\subsection{Qualitative Results}
\begin{figure*}[t]
    \centering
    \includegraphics[width=1.0\textwidth]{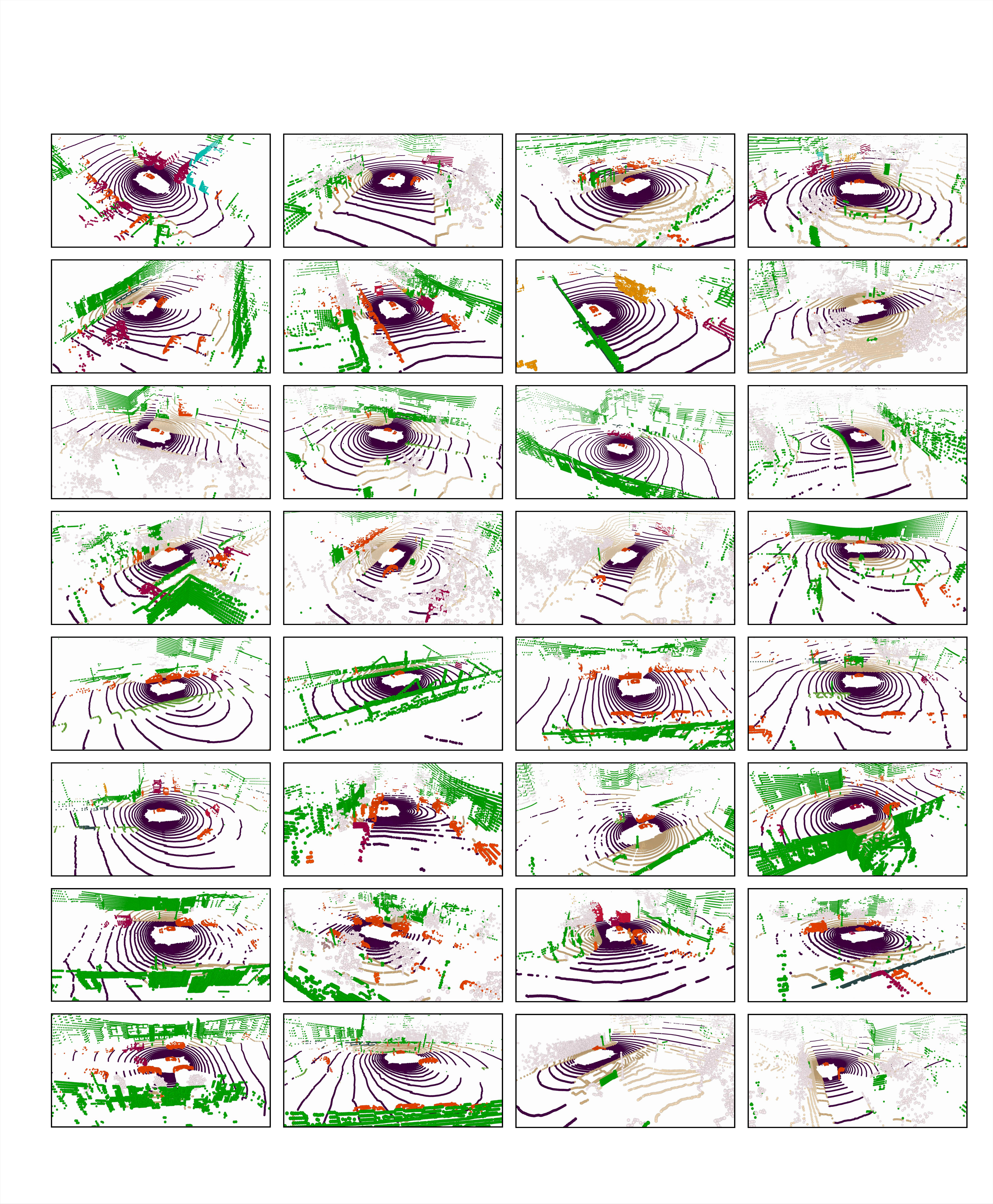}
    \caption{The qualitative results of Nuscenes test set.}
    \label{fig:test_nuscenes}
\end{figure*}

We provide visual qualitative results on SemanticKITTI dataset~\cite{behley2019semantickitti}, shown in Fig.~\ref{fig:test_kitti}. Furthermore, the visual results on Nuscenes~\cite{caesar2020nuscenes} is also provided in Fig.~\ref{fig:test_nuscenes}. Both results demonstrate the good semantic segmentation quality of our method.
\subsection{Failure Cases} 
\begin{figure*}[t]
    \centering
    \includegraphics[width=1.0\textwidth]{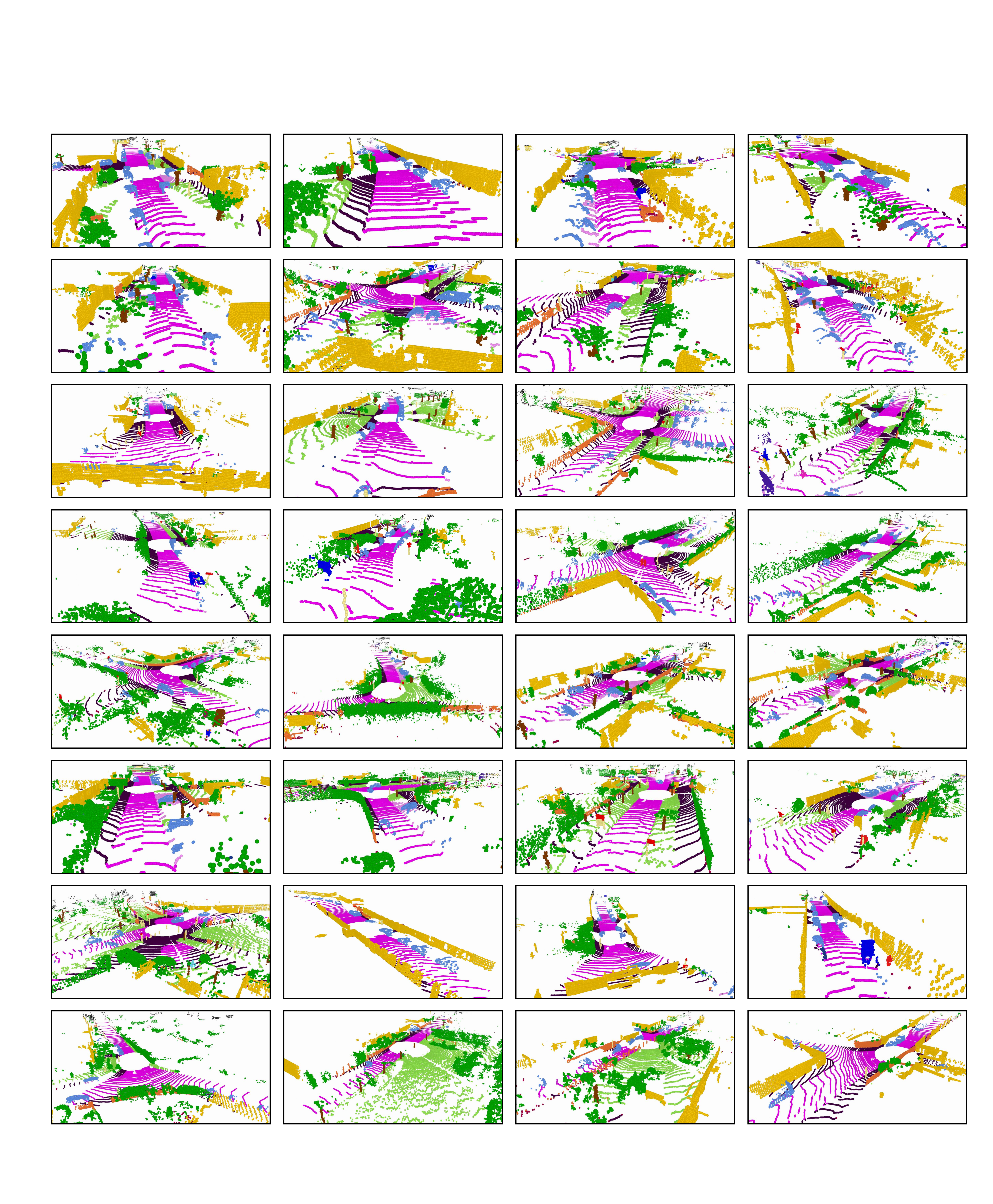}
    \caption{The qualitative results of SemanticKITTI test set.}
    \label{fig:test_kitti}
\end{figure*}

Although our GASN has achieved great improvements compared with previous methods, it still has some limitations and bad cases.
\begin{itemize}
    \item For those rare classes like motorcyclist, GASN has $0.\%$ performance in Tab.~\ref{tab:metricsKitti}, showing incapability to handle some extreme imbalanced data distribution problems.
    \item The predictions of the background categories like fence and building are not good enough. We can see some inconsistent predictions in the last image of Fig.~\ref{fig:test_kitti} due to lack of enough context information.
    \item Some over-segmentation and under-segmentation problems in the third row of Fig.~\ref{fig:test_kitti}. Therefore, we may cope with this problem by extending the fine-grained features learning.
\end{itemize}

\section{Sample codes}
\label{sec:sample_codes}
We also provide our model file implemented with PyTorch~\cite{paszke2019pytorch} in the attached file~\textbf{gasn.py}. For the implementation of our proposed module, we utilize some public repos, like torchscatter~\cite{pytorch_scatter} and spconv~\cite{spconv} for GPU acceleration.    

\bibliography{aaai22}